# Graph-Free Learning in Graph-Structured Data: A More Efficient and Accurate Spatiotemporal Learning Perspective

Xu Wang[1], Pengfei Gu[1], Pengkun Wang[1], Binwu Wang[1], Zhengyang Zhou[1], Lei Bai[2*], Yang Wang[1*]
[1]University of Science and Technology of China, Hefei, China
[2]Shanghai AI Laboratory, Shanghai, China

## ABSTRACT

Spatiotemporal learning, which aims at extracting spatiotemporal correlations from the collected spatiotemporal data, is a research hotspot in recent years. And considering the inherent graph structure of spatiotemporal data, recent works focus on capturing spatial dependencies by utilizing Graph Convolutional Networks (GCNs) to aggregate vertex features with the guidance of adjacency matrices. In this paper, with extensive and deep-going experiments, we comprehensively analyze existing spatiotemporal graph learning models and reveal that extracting adjacency matrices with carefully design strategies, which are viewed as the key of enhancing performance on graph learning, are largely ineffective. Meanwhile, based on these experiments, we also discover that the aggregation itself is more important than the way that how vertices are aggregated. With these preliminary, a novel efficient Graph-Free Spatial (GFS) learning module based on layer normalization for capturing spatial correlations in spatiotemporal graph learning. The proposed GFS module can be easily plugged into existing models for replacing all graph convolution components. Rigorous theoretical proof demonstrates that the time complexity of GFS is significantly better than that of graph convolution operation. Extensive experiments verify the superiority of GFS in both the perspectives of efficiency and learning effect in processing graph-structured data especially extreme large scale graph data.

## 1 INTRODUCTION

In recent years, massive amount of spatiotemporal data have been collected in various fields, e.g., urban computing, meteorology, and atmosphere quality. Such collected spatiotemporal data is a set of correlated time series where each individual sequence is about the temporal variation of the monitored information at a specific physical location. And Spatiotemporal learning, which aims at extracting spatiotemporal correlations from the collected spatiotemporal data, has attracted more and more attentions [2, 4, 25, 29].

Early efforts in spatiotemporal learning mostly focus on extracting both spatial and temporal correlations respectively with Convolution Neural Networks (CNNs) [31, 36, 41] and Recurrent Neural Networks (RNNs) [21, 24, 35]. However, Such CNN based methods, which divide the whole space into grids to extract spatial correlations, has never considered the irregular spatial distributions of spatiotemporal data. Therefore, this will definitely lead to the inevitable missing of topology information and non-Euclidean correlations in spatial learning. To tackle the above-mentioned issues of CNN based methods, recent spatiotemporal graph learning works [18, 22, 33] tend to model spatiotemporal data into graph structure by modelling spatial points as vertices. They first employ various strategies to extract adjacency matrices for comprehensively and precisely modeling the spatial correlations among vertices, and then use graph convolution to aggregate the features of vertices based on the extracted adjacency matrices.

Existing works on spatiotemporal graph learning can be divided into two categories, predefined adjacency matrix based methods and learnable adjacency matrix based methods. Regarding the first category, they incorporate predefined distance based [18, 33, 39] or temporal similarity based [11, 17, 20] adjacency matrices with graph convolution. However, such methods assume that vertices, which are in closer distances or with similar temporal series, are highly correlated, and the adjacency matrices are predefined before training and keep fixed during training. This determines that these methods cannot effectively extract the time-varying correlations among vertices due to their invariable adjacency matrices. Considering the lacking of the representation ability of these predefined adjacency matrix based methods, the second category focus on extracting adjacency matrices in a learnable manner during training [3, 27, 28] or throughout [14, 16]. The previous methods can effectively enhance their representation abilities in spatial perspective with their learnable adjacency matrices, and the learning of adjacency matrices are only within the training period. The latter methods can even represent dynamic spatial dependencies into dynamic adjacency matrices with their embedded self-attention mechanism.

Throughout all existing works on spatiotemporal graph learning, we discover that how to effectively extract spatial correlations is one of the core concerns of them. Therefore, a very simple intuition is that the performances of existing works improve with the increasing of the complexities of models, and this easily leads all researchers to an unsubstantiated conclusion: *The extraction of spatial adjacency matrix is critical to the success of such spatiotemporal learning approaches.* However, by comprehensively analyzing the evolution of existing works, we discover that they are usually improved in three aspects, i) new approach for extracting spatial adjacency matrix, ii) new method for extracting temporal dependencies, and iii) new mechanism for more comprehensive fusion between spatial and temporal correlations. Therefore, it is baseless to simply attribute the improvements of these models to the construction of their adjacency matrices. To explicitly verify this deduction, we first classify all existing methods into four disaggregated classifications, select STGCN [33], STFGNN [17], AGCRN [3], ASTGNN [16] as the four representative methods respectively, and carry out a series of experiments by replacing the learned adjacency matrices of the four representative approaches respectively with a random matrix, a matrix filled with a specific value, and some variant matrices by combining the previous two matrices with an identity matrix, and the result indicates **a first fact:** *the performances of such approaches*

---

Xu Wang[1], Pengfei Gu[1], Pengkun Wang[1], Binwu Wang[1], Zhengyang Zhou[1], Lei Bai[2*], Yang Wang[1*]

*can hardly be influenced by replacing their adjacency matrices with generated ones, and this indicates that using different complex strategies to enhance the extraction of adjacency matrices, which is both resource-consuming and time-consuming, seem to be unnecessary and cancelable.* Meanwhile, this observation brings another question, i.e., is graph convolution based data aggregation still useful in enhancing the final performance of spatiotemporal graph learning? To answer this question, we design another series of experiments to verify the impacts of graph convolution based data aggregation by replacing the adjacency matrices of those approaches with an identity matrix, and the result indicates **a second fact:** *graph convolution based data aggregation appears useful in enhancing spatiotemporal graph learning for most existing approaches except STGCN.* To figure out whether graph convolution is useful or not in STGCN, we detailedly analyze the architecture of STGCN and discover that it has a unique module, layer normalization, which doesn't exist in the other three representative methods. Intuitively, employ layer normalization in vertex dimension leads to aggregations of vertices due to its integrated mean and covariance calculation operations, therefore, we conduct an additional experiment by removing the layer normalization operation from vertex dimension, and the result indicate **a third fact:** *aggregating vertices in spatial perspective is essential and important for graph learning, and layer normalization is also be of the functionality of data aggregation.* So far, in summary, we can obtain a simple conclusion, i.e., the aggregation of neighboring vertices is crucial and effective for spatiotemporal graph learning, but the way of adjacency matrix based graph convolution aggregating vertices has little additional effect on spatiotemporal learning.

Based on such finding, we think that existing sophisticated adjacency matrix and graph convolution based approaches are improvable, and then propose a novel efficient Graph-Free Spatial (GFS) learning module based on layer normalization for capturing spatial correlations in spatiotemporal graph learning. Specifically, we first theoretically prove that the operation of layer normalization in vertex dimension, which can also be transferred to a graph-convolution-like form, is equivalent to a series of complex matrix multiplications which are the core operation of data aggregation in the operation of graph convolution. Next, we propose a novel GFS learning module which consists of only a linear projection and ReLU activation function combined component and a layer normalization module. Rigorous theoretical proof demonstrates that the time complexity of GFS is significantly better than that of graph convolution operation. The proposed GFS module can be easily plugged into existing models for replacing all graph convolution components, as far as the dimensions are aligned. Extensive experiments verify the superiority of GFS in both the perspectives of efficiency and learning effectiveness.

The contributions of this paper can be summarized as follows,

- To the best of our knowledge, this paper first reveals the fact that adjacency matrix plays unimportant role in learning spatial correlations from graph structured data, and also for the fist time, this paper confirms that aggregating vertices in spatial perspective is essential and important for graph learning.

- We cross-verify that layer normalization is effective in aggregating data among vertices in both theoretical and experimental perspectives. Based on this, we design a novel efficient GFS learning module based on layer normalization for capturing spatial correlations in spatiotemporal graph learning. The proposed GFS module can be plugged into existing spatiotemporal graph learning models as an alternative to graph convolution layer.
- We conduct extensive experiments on several widely-used real-world graph-structured spatiotemporal datasets as well as two very large-scale graph datasets. To evaluate the effectiveness and efficiency of GFS module, we select a series of baselines and use GFS to replace their embedded graph convolution layer. Experimental result shows that GFS module is superior to traditional graph convolution in terms of both efficiency and learning effect.

The paper is organized as follows. In Section 2, we analyze and conclude related works, and then re-investigate existing spatiotemporal graph learning works with extensive experiments in Section 3. Based on all learned facts, in Section 4, we introduce the design and implementation of GFS learning module. Section 5 describes the experiments for evaluating our propose module and Section 6 concludes this paper.

## 2 RELATED WORKS

Spatiotemporal learning has attracted extensive research attentions [12, 19, 22, 26, 37] in recent years. Early works [15, 24, 31, 32, 34–36, 41] mainly focus on employing CNN based networks and RNN based networks to respectively extract both spatial and temporal correlations. Nevertheless, in spatial perspective, such CNN-based spatiotemporal learning methods mesh road network into regular grids and employ convolutional neural networks to capture spatial dependencies among grids. And those grid-partition based methods are incapable of capturing the inherent Non-Euclidean structured characteristics, hence are short in fully capture spatial correlations.

To this end, recent works, i.e., spatiotemporal graph learning, aim at modeling such these Non-Euclidean characteristics with graph structure [7, 13, 18, 38–40], and existing efforts on this branch can be roughly divided into two categories, predefined adjacency matrix based methods and learnable adjacency matrix based methods.

Regarding the first category, in spatial perspective, they incorporate predefined distance based [18, 33, 39] or temporal similarity based [11, 17, 20] adjacency matrices with graph convolution. Specifically, STGCN [33] applies graph convolution on fixed distance-based adjacency matrix, and convolutional network for modeling temporal dependencies. DCRNN [18] captures spatial dependencies with bidirectional random walks on graph, and captures temporal dependencies with a encoder-decoder framework and scheduled sampling. T-GCN [39] combines GCN with GRU to exploit the spatiotemporal correlations of urban traffics. On the other hand, STAG-GCN [20] applies classic DTW algorithm [5] to construct adjacency matrices based on the similarities of vertices temporal trends and combines temporal aware adjacency matrices with distance based adjacency matrices to enhance the modeling of spatial correlations. STFGNN [17] modifies DTW to address time efficiency concerns and involves temporal connections in the construction of adjacency matrices. However, such methods assume that vertices, which are in closer distances or with similar temporal series, are highly correlated, and the adjacency matrices are predefined before training and keep fixed during training. This determines that these methods cannot effectively extract the time-varying correlations among vertices due to their invariable adjacency matrices.

Given the fact that all above-mentioned predefined adjacency matrix based methods are poor at representing the time-varying correlations, recent works in spatiotemporal graph learning mostly construct adjacency matrices in a learnable manner during training [3, 28] or throughout [14, 16]. In particular, GraphWaveNet [28] maintains learnable embeddings for each vertex, and constructs adjacency matrices by calculating the embedded similarities between vertices. AGCRN [3] follows the idea of GraphWaveNet and further utilizes the embedded similarities between vertices to modify graph convolution for learning node-specific patterns. Moreover, ASTGCN [14] utilizes attention mechanism in both spatial and temporal dimension to capture the dynamic spatiotemporal correlations at different time intervals. By multiplying the attention values with distance based adjacency matrices, ASTGCN actually gets dynamic adjacency matrix changing at different time steps. ASTGNN [16] is an upgraded version of ASTGCN, which modifies the attention mechanism in ASTGCN and involves short-term temporal trends when calculating attention values.

As analyzed, existing works in spatiotemporal graph learning have devoted themselves into enhancing the effect of graph convolution. by generating various adjacency matrices with different strategies. Indeed, with the increasing of the complexity of adjacency matrix construction, the performances of such spatiotemporal graph learning methods increase correspondingly. This gives us an illusion that the performance enhancement mainly due to the improvement in adjacency matrix construction, and this is why researchers never get tired of designing new strategy to improve the adjacency matrix of graph convolution. However, by comprehensively analyzing the evolution of existing works, we discover that they are usually improved in three aspects, i) new approach for extracting spatial adjacency matrix, ii) new method for extracting temporal dependencies, and iii) new mechanism for more comprehensive fusion between spatial and temporal correlations. So, there naturally raises a question: how much can the modification of adjacency matrix helps on the final performance of spatiotemporal graph learning?

## 3 RE-INVESTIGATION ON EXISTING SPATIOTEMPORAL GRAPH LEARNING

To answer the above-proposed question, in this section, we first select some representative spatiotemporal graph learning methods based on the above analysis on spatiotemporal graph learning, and then design a series of experiment to comprehensively analyze them.

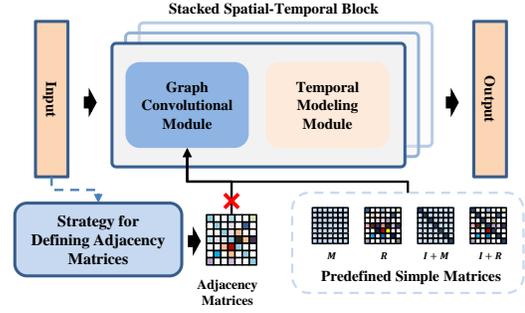

Figure 1: Illustration of replacing adjacency matrices of existing works.

### 3.1 Selection of representative spatiotemporal graph learning methods

In previous section, existing spatiotemporal graph learning works have been roughly divided into two major categories. Here, to specifically distinguish and evaluate the impacts of adjacency matrix, we further classify all existing works into four subclasses:

- **Methods with distance based predefined adjacency matrix:** The adjacency between two vertices is determined based on the physical distance between them, and we here select STGCN [33] as the representative method of this subclass.
- **Methods with temporal similarity based predefined adjacency matrix:** The adjacency between two vertices is determined based on the similarity of the temporal patterns of these two vertices, and select STFGNN [17] as the representitive method of this subclass.
- **Methods with adjacency matrix learned during training:** The adjacency matrix is determined based on the similarities of embeddings which are learned from vertices during training period, such adjacency matrices are expected to model latent correlations among vertices. We here select AGCRN [3] as the representative model of this subclass.
- **Methods with dynamic adjacency matrix:** The adjacency matrix, which is constructed dynamically based on attention or some other mechanisms, are data-driven generated and changes over time.

### 3.2 Re-investigation on the impacts of adjacency matrix

Regarding the selected four representative works, to re-investigate the impacts of adjacency matrix on their performances, we conduct a series of experiments by replacing their adjacency matrices with some artificially designed adjacency matrices, and the detailed operation of replacing adjacency matrices of existing works is demonstrated in Figure 5. The artificially generated adjacency matrices are,

- **R**: The matrix filled with random values. Using **R** as adjacency matrix results in randomly aggregating of all vertices. Here **R** is generated by

$$\mathbf{R} = \mathrm{softmax}(\mathbf{R'}) \qquad (1)$$

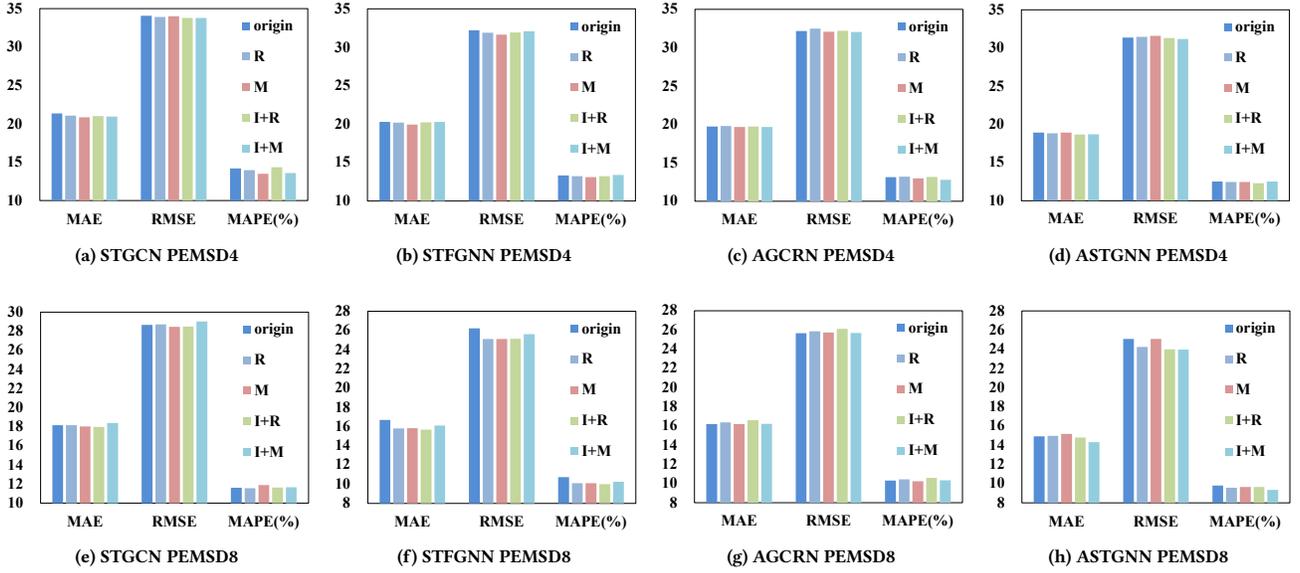

Figure 2: Performances of four representative methods with generated adjacency matrices on PEMSD4 and PEMSD8.

where
$$\mathbf{R}' = (r_{ij})_{N \times N} \quad (2)$$

Notice that $\mathbf{R}'_{ij} \sim \mathcal{N}(0, 1)$ which indicates that all elements of $\mathbf{R}'$ are sampled from standard Gaussian distribution, and $N$ corresponds to the number of vertices.

- **M**: Matrix filled with a specific value $1/N$. Using **M** as adjacency matrix results in equally aggregating of all vertices.
- **I + R**: Add self-loop on matrix **R**. Here **I** corresponds an identity matrix which is filled with 1.
- **I + M**: Add self-loop on matrix **M**. Here **I** corresponds an identity matrix which is filled with 1.

Notice that all these four alternative matrices are all $N \times N$ dimensional matrices. By comparing the original performances of the four representative methods and their performances in case that their adjacency matrices are respectively replaced with **R**, **M**, **I + R**, and **I + M**, we can investigate the impacts of different adjacency matrices on them, and the detailed implementation how the adjacency matrices are replaced are listed as follows,

- **STGCN**: We utilizes the publicly available implementation of STGCN [1], and simply replace the adjacency matrix calculated by STGCN with the generated matrices.
- **STFGNN**: The official implementation of STFGNN is utilized [2]. STFGNN proposes a spatial-temporal fusion graph which is constructed by combining both distance based and temporal similarity based adjacency matrices. Here both two matrices are replaced with our proposed matrices and the construction procedure of spatial-temporal fusion graph is kept.
- **AGCRN**: The adjacency matrix of AGCRN is determined based on the similarity of vertex embeddings. We sidestep the calculation of similarity and directly utilize the generated matrices as the adjacency matrix of AGCRN. All other settings of AGCRN are retained. The official implementation [3] is used.
- **ASTGNN**: A modified self-attention module is proposed in ASTGNN, which generates attention weights among vertices. The final adjacency matrix of ASTGNN is the dot-product of attention weights and the distance-based adjacency matrix. Therefore, we replace the final adjacency matrix of ASTGNN with our generated matrices. Similarly, the official implementation of ASTGNN is used [4].

The used datasets, metrics, and task settings are as follows,

- **Datasets:** All the experiments are conducted on PEMSD4 and PEMSD8 [6]. More details about the datasets and data preprocessing will be introduced in Section "EXPERIMENTS".
- **Task:** The task is to predict the spatiotemporal data over the next 12 time steps based on the data over the past 12 time steps.
- **Metrics:** Three widely used evaluation metrics are employed to measure the prediction accuracy. We here detail the definitions of all the three metrics. Let $V \in \mathbb{R}^{T \times N \times 1}$ denote the ground truth future data of all $N$ vertices during $T$ time steps and $\hat{V} \in \mathbb{R}^{T \times N \times 1}$ denote the predicted values. The metrics can be formulated as follows.

---
[1] https://github.com/hazdzz/STGCN
[2] https://github.com/MengzhangLI/STFGNN
[3] https://github.com/LeiBAI/AGCRN
[4] https://github.com/guoshnBJTU/ASTGNN

$$\text{MAE}(V, \hat{V}) = \frac{1}{TN} \sum_{i=1}^{T} \sum_{j=1}^{N} |V_{ij} - \hat{V}_{ij}| \quad (3)$$

$$\text{RMSE}(V, \hat{V}) = \sqrt{\frac{1}{TN} \sum_{i=1}^{T} \sum_{j=1}^{N} (V_{ij} - \hat{V}_{ij})^2} \quad (4)$$

$$\text{MAPE}(V, \hat{V}) = \frac{1}{TN} \sum_{i=1}^{T} \sum_{j=1}^{N} \left|\frac{V_{ij} - \hat{V}_{ij}}{V_{ij}}\right| \quad (5)$$

By replacing the adjacency matrices of the four representative methods with the generated matrices, **R**, **M**, **I + R**, and **I + M**, we can evaluate the effects and validity of the extracted adjacency matrices of the four representative methods, and the results are demonstrated in Figure 2. As demonstrated, while the extracted adjacency matrices of these four methods are replaced by **R**, **M**, **I + R**, and **I + M**, respectively, the performances of them barely changed with two different datasets. Specifically, while the extracted adjacency matrices are replaced by **R**, **M**, **I + R**, and **I + M**, on PEMSD4, the performances of STGCN, STFGNN, AGCRN, ASTGNN change by {0.41%, 1.06%, 2.14%} at worst and {2.54%, 1.79%, 2.82%} in average in terms of MAE, RMSE and MAPE respectively, and change respectively by {1.18%, 0.65%, 0.87%} at worst and {1.20%, 1.50%, 1.63%} in average on PEMSD8. This reveals a very important fact: *the effect of extracting an adjacency matrix with carefully designed strategy in spatiotemporal graph learning is limited and even non-existent, and the complex and time-consuming strategies for extracting and constructing adjacency matrices within those existing spatiotemporal learning efforts seem to be unnecessary and cancelable.* On the other hand, considering the effect of enhancing adjacency matrix is limited, is graph convolution based data aggregation still useful in enhancing the final performance of spatiotemporal graph learning?

### 3.3 Investigation on the impacts of graph convolution based data aggregation

To answer the question proposed in the previous subsection, in this subsection, we design another series of experiments to evaluate the impacts of graph convolution based data aggregation. Specifically, regarding the four representative spatiotemporal graph learning methods, we use the previous defined identity matrix **I** to make sure that the feature of each vertex is isolated and will never been aggregated with the feature of any other vertices. The datasets, task, metrics and all other basic settings in this series of experiments are the same to those in the experiments in the previous subsection, and the results are reported in Table 1. Notice here the column with *Origin* corresponds the performances of the four models with their original adjacency matrices and the column with **I** means the performances of them while their adjacency matrices were replaced with **I**. As reported in this table, once the adjacency matrices were replaced with **I**, on both PEMSD4 and PEMSD8, the performances of STFGNN, AGCRN and ASTGNN decrease significantly in terms of all metrics. Specifically, when the adjacency matrices are replaced with **I**, on PEMSD4, the performances of STFGNN, AGCRN and ASTGNN decrease by {17.25%, 15.32%, 17.48%}, {18.95%, 16.23%, 18.80%} and {19.07%, 17.06%, 20.61%} in terms of MAE, RMSE and MAPE, and {9.11%, 10.64%, 7.09%}, {16.93%, 16.10%, 17.28%} and

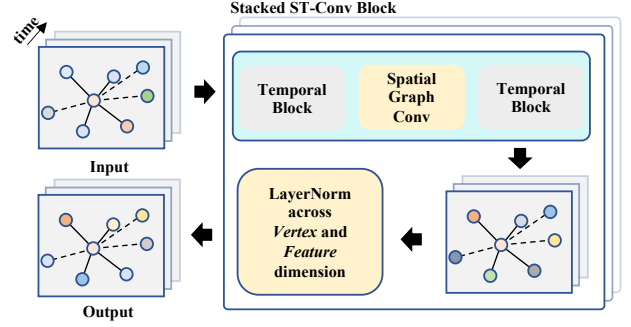

Figure 3: Architecture of STGCN.

{12.52%, 12.44%, 11.42%} on PEMSD8. *This can easily lead us to the conclusion that graph convolution based data aggregation are still useful in enhancing spatiotemporal graph learning.* However, on the other hand, we discover that the performances of STGCN are barely change whether its adjacency matrix is replaced or not. This seems counter the earlier conclusion that we just obtained from the experiments on STFGNN, AGCRN and ASTGNN. Therefore, here a big confusion comes naturally, whether graph convolution based data aggregation is effective or not in enhancing spatiotemporal learning?

Table 1: Performances of four representative methods with I on PEMSD4 and PEMSD8.

| Datasets | Metrics | STGCN | | STFGNN | | AGCRN | | ASTGNN | |
|---|---|---|---|---|---|---|---|---|---|
| | | Origin | I | Origin | I | Origin | I | Origin | I |
| PEMSD4 | MAE | 21.34 | 21.67 | 20.29 | 23.79 | 19.74 | 23.48 | 18.93 | 22.54 |
| | RMSE | 34.07 | 34.47 | 32.23 | 37.17 | 32.16 | 37.38 | 31.36 | 36.71 |
| | MAPE(%) | 14.18 | 14.05 | 13.33 | 15.66 | 13.14 | 15.61 | 12.52 | 15.10 |
| PEMSD8 | MAE | 18.15 | 19.17 | 16.68 | 18.20 | 16.18 | 18.92 | 14.94 | 16.81 |
| | RMSE | 28.67 | 29.52 | 26.23 | 29.02 | 25.65 | 29.78 | 25.08 | 28.20 |
| | MAPE(%) | 11.61 | 12.18 | 10.72 | 11.48 | 10.30 | 12.08 | 9.81 | 10.93 |

### 3.4 Further investigation on STGCN

In the previous subsection, we discover an interesting issue that graph convolution based data aggregation seems essential in STFGNN, AGCRN and ASTGNN, while it is almost useless in STGCN, and this greatly encourages us to further investigate STGCN. As illustrated in Figure 3, we discover that STGCN includes an additional layer normalization [1] module which doesn't exist in the other three representative methods. Specifically, STGCN applies layer normalization on both vertices dimension and feature dimension. Given input of all the features of all vertices at a specific time point, $\mathbf{V} \in \mathbb{R}^{N \times D}$, the layer normalization can be briefly formulated as,

$$\text{LN}(\mathbf{V}) = \frac{\mathbf{V} - \text{E}[\mathbf{V}]}{\sqrt{\text{Var}[\mathbf{V}]}} \bigotimes \mathbf{W} + \mathbf{B} \quad (6)$$

$\mathbf{W}, \mathbf{B} \in \mathbb{R}^{N \times D}$ are learnable affine parameters and $\bigotimes$ is element-wise multiplication. $\text{E}[\mathbf{V}]$ and $\text{Var}[\mathbf{V}]$ are the mean and covariance of **V** respectively. Due to the operations of computing mean and covariance, applying layer normalization on vertex dimension implicitly leads to aggregation of all vertices. Therefore, this module

Xu Wang[1], Pengfei Gu[1], Pengkun Wang[1], Binwu Wang[1], Zhengyang Zhou[1], Lei Bai[2*], Yang Wang[1*]

is also of the functionality of data aggregation. So far, we conject that whether the data aggregation among vertices is what really works? To clarify such conjecture, we design an additional experiment. Here we directly modify STGCN by only employing layer normalization on feature dimension and removing the implicit aggregation of vertices, and the result are reported in Table 2. As shown in this table, in case that the data aggregation operation is removed from STGCN, the performances of STGCN* are significantly worse than those of STGCN. In particular, in case that the STGCN uses its original extracted adjacency matrix, comparing with the performances of STGCN, the performances of STGCN* decreases by {50.89%, 42.68%, 62.41%} on PEMSD4 in terms of MAE, RMSE and MAPE, and decrease by {29.92%, 25.18%, 25.32%} respectively on PEMSD8. And while the adjacency matrix is replaced with **I**, the performances of STGCN* further decrease by {19.21%, 17.12%, 21.75%} on PEMSD4 and {28.50%, 24.77%, 32.58%} on PEMSD8 respectively. In summary, *these experiments explicitly and undoubtedly verify the effectiveness and importance of aggregating vertices in spatial learning, and simultaneously indicates that layer normalization is also be of the functionality of data aggregation.*

Table 2: Impacts of data aggregation operation in STGCN.

| Datasets | Metrics | STGCN | | STGCN* | |
| --- | --- | --- | --- | --- | --- |
| | | Origin | I | Origin | I |
| PEMSD4 | MAE | 21.34 | 21.67 | 31.02 | 36.98 |
| | RMSE | 34.07 | 34.47 | 46.79 | 54.80 |
| | MAPE(%) | 14.18 | 14.05 | 21.79 | 26.53 |
| PEMSD8 | MAE | 18.15 | 19.17 | 23.58 | 30.30 |
| | RMSE | 28.67 | 29.52 | 35.89 | 44.78 |
| | MAPE(%) | 11.61 | 12.18 | 14.55 | 19.29 |

### 3.5 Lessons learned in re-investigating existing spatiotemporal graph learning methods

In this subsection, we will briefly summarize the lessons learned in re-investigating existing spatiotemporal graph learning with three series of carefully designed experiments.

- **LL1.** The core idea of existing spatiotemporal graph learning methods, i.e., using different complex strategies to enhance the extraction of adjacency matrices, which is both resource-consuming and time-consuming, is noneffective to the performance of spatiotemporal graph learning.
- **LL2.** Data aggregation among vertices, which can effectively capture spatial correlations among vertices, is the key to the success of spatiotemporal graph learning. Even with a matrix with a fixed value or random values, traditional graph convolution is still an effective way to achieve data aggregation, however it time complexity is also an issue that should be concerned about.
- **LL3.** Layer normalization, which is also be of the functionality of data aggregation, can also significantly enhance the performance of spatiotemporal graph learning by effectively extracting the spatial correlations among vertices.

And compared with traditional graph convolution, the time complexity of layer normalization has obvious advantages.

## 4 GRAPH-FREE SPATIAL MODULE IN SPATIOTEMPORAL GRAPH LEARNING

In this section, based on the previous learned lessons, we analyze the possible technical solution of spatiotemporal graph learning, and propose a graph-free spatial learning module as an alternative to graph convolution.

### 4.1 Rethinking of spatial learning in spatiotemporal graph learning

Based on the learned lessons **LL1** and **LL2**, we discover that graph convolution based spatial learning is highly inefficient. Specifically, the strategies for defining better adjacency matrices contribute little to the final effect of spatiotemporal graph learning, and though, by replacing the learned adjacency matrix with the fore-mentioned matrix **M**, the efficiency of traditional graph convolution can significantly improved on the premise of without significantly losing the performance of spatiotemporal graph learning, the computation complexity of graph convolution itself is also a serious issue that should be concerned about. Therefore, regarding spatial learning in spatiotemporal graph learning, we should pay more attentions to *aggregating vertices in a more efficient manner* rather than seeking strategies for defining adjacency matrices.

On the other hand, the learned lesson **LL3** indicates that layer normalization across vertex dimension is also valid to data aggregation, however, in theory, whether or why layer normalization has the additional functionality of data aggregation still needs to be further explored and analyzed.

### 4.2 Theoretical analysis on the effect of layer normalization in spatial learning

As analyzed in Section 3.4, layer normalization has shown its partial effectiveness on capturing spatial dependencies among vertices. Here, we further explore the data aggregation operation in layer normalization and analyze the correlations between layer normalization and graph convolution. The operation of layer normalization is defined in Equation 6. To simplify the following analysis, we set the feature dimension as 1 and omit the bias term **B** in layer normalization. Also, the temporal dimension is omitted since it is not relevant for the calculation. Based on the simplification, we have input $\mathcal{V} \in \mathbb{R}^{N \times 1}$, and layer normalization first calculates the mean of $\mathcal{V}$, i.e.,

$$\mathrm{E}[\mathcal{V}] = \mathfrak{R} \times \mathcal{V} \quad (7)$$

where $\mathfrak{R} \in \mathbb{R}^{N \times N}$ is a matrix filled with $\frac{1}{N}$. Thus, the term $\mathcal{V} - \mathrm{E}[\mathcal{V}]$ in Equation 6 can be calculated by,

$$\mathcal{V} - \mathrm{E}[\mathcal{V}] = (\mathbf{I} - \mathfrak{R}) \times \mathcal{V} \quad (8)$$

$\mathbf{I} \in \mathbb{R}^{N \times N}$ corresponds to the identity matrix. Considering the affine parameter $\mathbf{W} \in \mathbb{R}^{N \times 1}$ in Equation 6, the element-wise multiplication between $\mathcal{V} - \mathrm{E}[\mathcal{V}]$ and $\mathbf{W}$ can be transferred to a matrix multiplication, i.e.,

$$(\mathcal{V} - \mathrm{E}[\mathcal{V}]) \bigotimes \mathbf{W} = \mathrm{Diag}(\mathbf{W}) \times (\mathbf{I} - \mathfrak{R}) \times \mathcal{V} \quad (9)$$

where Diag(**W**) is a diagonal matrix whose diagonal element in each row corresponds to the element in the corresponding row of **W**, and $\bigotimes$ is element-wise product. Finally, by defining $\sigma_\mathcal{V} = \text{Var}[\mathcal{V}]$ which represents the standard deviation of $\mathcal{V}$, the calculation of layer normalization can be written as,

$$\text{LN}(\mathcal{V}) = [\frac{\text{Diag}(\mathbf{W})}{\sigma_\mathcal{V}}(\mathbf{I} - \mathfrak{R})]\mathcal{V} = \mathcal{A} \times \mathcal{V} \quad (10)$$

Where $\mathcal{A} = [\frac{\text{Diag}(\mathbf{W})}{\sigma_\mathcal{V}}(\mathbf{I} - \mathfrak{R})]$. Regarding the calculation of layer normalization, the diagonal matrix Diag(**W**) ensures the diversity among the values in different rows in $\mathcal{A}$, and $\mathcal{A}$ contains data-driven term $\sigma_\mathcal{V}$ and learnable parameters **W**, which determine that layer normalization is effective and flexible in aggregating vertices. In case that the dimensionality of feature is $D$, we should calculate each dimension individually with the above-mentioned method. Figure 4 illustrates the layer normalization calculation for the $i$-th dimension where $\mathcal{V}_i$ the $i$-th dimensional feature.

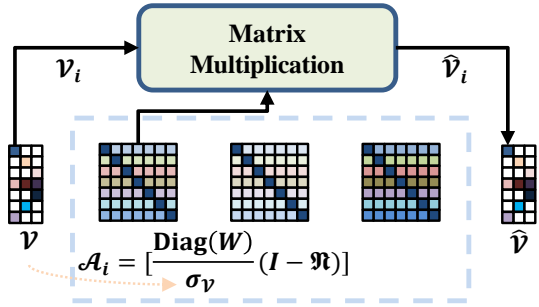

**Figure 4: Graph-convolution-like form of layer normalization in data aggregation among vertices.**

Comparing layer normalization with the calculation of graph convolution defined as,

$$\text{GraphConv}(\mathcal{V}, \mathbf{A}) = \mathbf{A} \times \mathcal{V} \times \mathbf{W}' \quad (11)$$

where **A** and **W**′ correspond to the adjacency matrix and the learnable parameters respectively, we discover that layer normalization aggregates vertices in a way similar to graph convolution. Actually, the aggregation of vertices in layer normalization can be seen as a graph convolution layer equipped with adjacency matrix as $\mathcal{A}$ in Equation 10 and the difference between the two is that layer normalization has no linear projection on the feature dimension, i.e., **W**′ in Equation 11.

So far, we theoretically analyze the effectiveness of layer normalization on data aggregating and prove that the operation of layer normalization is similar to graph convolution without linear projection on feature dimension. Considering the superiority of layer normalization in terms of computational complexity, the idea that how to achieve a more efficient spatial learning module by taking advantage of layer normalization is worth exploring.

### 4.3 Layer normalization based graph-free spatial learning

Based on the analysis in the previous subsection, we propose a novel and efficient Graph-Free Spatial (GFS) learning module by equipping layer normalization with some add-on components. The detailed architecture of GFS learning module is illustrated in Figure 5. As shown, we equip layer normalization with a linear projection and ReLU activation function to extract spatial correlations among vertices. Considering that residual connection is effective to increase the representation power of graph convolution [10], we also employ a residual connection in GFS learning to increase the representation ability of layer normalization.

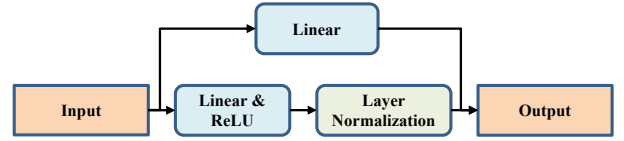

**Figure 5: Architecture of layer normalization based graph-free spatial learning module.**

Specifically, given input $\mathcal{V} \in \mathbb{R}^{Q \times N \times d_{in}}$, GFS first applies a linear projection and a ReLU activation function on feature dimension, i.e.,

$$\mathcal{V}' = \text{ReLU}(\mathcal{V}\mathbf{W_{s1}} + \mathbf{b_{s1}}) \quad (12)$$

where $\mathbf{W_{s1}} \in \mathbb{R}^{d_{in} \times d_{out}}$ and $\mathbf{b_{s1}} \in \mathbb{R}^{d_{out}}$ are learnable parameters, and $\mathcal{V}' \in \mathbb{R}^{Q \times N \times d_{out}}$ is the result of projection. Next, a layer normalization is applied on $\mathcal{V}'$ on both vertex and feature dimensions, i.e.,

$$\hat{\mathcal{V}} = \frac{\mathcal{V}' - \text{E}[\mathcal{V}']}{\sqrt{\text{Var}[\mathcal{V}']}} \bigotimes \mathbf{W_{s2}} + \mathbf{B_{s2}} \quad (13)$$

where $\mathbf{W_{s2}} \in \mathbb{R}^{N \times d_{out}}$ and $\mathbf{B_{s2}} \in \mathbb{R}^{N \times d_{out}}$ are learnable affine parameters, $\hat{\mathcal{V}}$ is the output of layer normalization. To generate the final output $\mathcal{V}_{out} \in \mathbb{R}^{Q \times N \times d_{out}}$ of GFS, a residual connection integrated with linear projection is applied on the input and added to $\hat{\mathcal{V}}$, i.e.,

$$\mathcal{V}_{res} = \text{ReLU}(\mathcal{V}\mathbf{W_{res}} + \mathbf{b_{res}}) \quad (14)$$
$$\mathcal{V}_{out} = \mathcal{V}_{res} + \hat{\mathcal{V}} \quad (15)$$

where $\mathbf{W_{res}} \in \mathbb{R}^{d_{in} \times d_{out}}$ and $\mathbf{b_{res}} \in \mathbb{R}^{d_{out}}$ are learned parameters. Notice that GFS module can be easily plugged into existing models for replacing all graph convolution components, as far as the dimensions are aligned.

### 4.4 Time complexity comparison between GFS and graph convolution

To validate the efficiency of GFS theoretically, in this subsection, we compare the time complexity of GFS with that of standard graph convolution defined in Equation 11.

- **Time complexity of GFS**: As illustrated in Figure 5, GFS contains two linear projection operations on feature dimension, and each linear projection operation has the time complexity of $O(N \times d_{in} \times d_{out})$. Regarding the operation of layer normalization, its embedded mean calculation, standard deviation calculation, and element-wise multiplication are all in linear time complexity in both the dimensionalities of vertex and feature, i.e., $O(N \times d_{out})$. Therefore, the overall time complexity of GFS, which is

Xu Wang[1], Pengfei Gu[1], Pengkun Wang[1], Binwu Wang[1], Zhengyang Zhou[1], Lei Bai[2*], Yang Wang[1*]

$O(N \times d_{out} + N \times d_{in} \times d_{out})$, is exactly linear to the number of vertices.
- **Time complexity of standard graph convolution**: As defined in Equation 11, graph convolution contains two matrix multiplications. The first multiplication between adjacency matrix and features of vertices has the time complexity of $O(N^2 \times d_{in})$, and The latter multiplication with learnable parameters has the same time complexity with linear projection in GFS, i.e., $O(N \times d_{in} \times d_{out})$. Therefore, the overall time complexity of graph convolution is $O(N^2 \times d_{in} + N \times d_{in} \times d_{out})$, which is quadratic to the number of vertices.

As theoretically analyzed, the proposed GFS is more efficient than the operation of graph convolution, and we reckon that such difference in efficiency will be reflected most vividly in processing very large scale graph.

## 5 EXPERIMENTS
### 5.1 Experimental scheme and datasets

To evaluate the performance of our proposed GFS module, we select a series of representative graph convolution based spatiotemporal graph learning works and replace their integrated graph convolution module with GFS module. The experiments consist of three parts: **i) Performances on spatiotemporal graph learning:** Based on four widely-used real-world spatiotemporal datasets, PEMSD3, PEMSD4, PEMSD7 and PEMSD8 [6][5], we conduct a series of experiments to compare the performances of GFS and graph convolution on different backbones in terms of traffic prediction, **ii) Performances on graph learning with extreme large graph:** Based on two graph datasets with extreme large numbers of vertices, i.e., PubMed [30] and Coauthor Physics [23], regarding the task of node classification, we further investigate the time efficiency and capability of GFS in modeling spatial correlations, and **iii) Investigation on graph-free architecture:** To further verify the effectiveness of each individual component of the graph-free architecture, we carry out a series of ablative studies on PeMSD4. The statistical information of all used datasets is summarized in Table 3.

Table 3: Dataset descriptions.

| Dataset | #Vertices | #Features | Time Range |
| --- | --- | --- | --- |
| PeMSD3 | 358 | 1 | 09/01/2018 - 11/30/2018 |
| PeMSD4 | 307 | 1 | 01/01/2018 - 02/28/2018 |
| PeMSD7 | 883 | 1 | 05/01/2017 - 08/31/2017 |
| PeMSD8 | 170 | 1 | 07/01/2016 - 08/31/2016 |
| PubMed | 19717 | 500 | N/A |
| Coauthor Physics | 495924 | 8415 | N/A |

### 5.2 Data preprocess

For spatiotemporal datasets, linear interpolation is utilized to fill the missing values in the datasets. Then, we apply min-max normalization to normalize all data into the range of $[-1, 1]$ to stabilize the training process. Regarding all experiments on spatiotemporal graph learning, all spatiotemporal datasets are divided into training, validation and testing sets with the ratio of 6:2:2 in chronological order, i.e., the earliest 60% are used for training, the subsequent 20% are used for validation, and the last samples are for testing. Notice that the raw traffic flow data within spatiotemporal datasets is aggregated with the interval of 5 minutes, therefore the aggregated datasets contain 288 data points for each day. For graph datasets, we directly use the data provided by torch_geometric [6] and split PubMed and Coauthor Physics with the ratio of 9:1 and 7:3 respectively for training and testing.

### 5.3 Backbones and experimental settings

**Backbones**: to compare the performances of GFS and graph convolution, we select a series of graph convolution based backbones including:

- **STGCN** [33]: deploys graph convolution and temporal convolution for capturing spatial and temporal dependencies, respectively.
- **DCRNN** [18]: combines diffusion graph convolution with recurrent units for multi-step prediction.
- **GraphWaveNet** [28]: proposes node embeddings for constructing adjacency matrices and combines GCN with dilated casual convolution for traffic forecasting.
- **ASTGCN** [14]: is a self-attentive traffic forecasting model, and captures the dynamics in a flexible manner.
- **AGCRN** [3]: proposes node-adaptive graph convolution, generates node-specific parameters according to learnable node embeddings, and combines it with GRU [8].
- **STFGNN** [17]: constructs temporal graphs based on the similarities between time series of vertices by utilizing DTW algorithm. The temporal graphs are fused with distance-based graphs for better modeling spatial dependencies.
- **STG-NCDE** [9]: extends the concept of neural controlled differential equations and designs two novel NCDEs for spatial and temporal processing, respectively.
- **ASTGNN** [16]: is an upgraded version of ASTGCN by modifying the attention mechanism in ASTGCN and adding positional embedding into the model.

**Experimental settings**: Regarding the experiments of spatiotemporal forecasting, to evaluate the effect of GFS module, we replace the graph convolution component of all backbones with our proposed GFS module and keep all other settings of those backbones unchanged. The metric of MAE are chosen as the loss, and two more metrics, RMSE and MAPE, are additionally evolved to comprehensively evaluate all models. In case that GFS is employed on existing models, we strictly follow the training settings of the original models for fair comparison, including optimizers, batch size, maximum epochs, etc. Regarding the learning on extreme large graphs, a stack of three layer GFSs and a stack of three layer GCNs are trained on randomly selected training samples with NLLLoss [7]. Notice that all experiments are executed with one E5-2620 v4 @ 2.10GHz CPU and one Nvidia Tesla V100 16GB GPU.

---

[5]These four datasets, which are about the highway traffic flow in California, are collected by Caltrans Performance Measurement System.

[6]https://pytorch-geometric.readthedocs.io/en/latest/modules/datasets.html
[7]https://pytorch.org/docs/stable/generated/torch.nn.NLLLoss.html

Graph-Free Learning in Graph-Structured Data: A More Efficient and Accurate Spatiotemporal Learning Perspective

Table 4: Performance comparison between backbones and their GFS variants. # denotes that GFS is integrated.

| Model | PEMSD3 | | | PEMSD4 | | | PEMSD7 | | | PEMSD8 | | |
| --- | --- | --- | --- | --- | --- | --- | --- | --- | --- | --- | --- | --- |
| | MAE | RMSE | MAPE(%) | MAE | RMSE | MAPE(%) | MAE | RMSE | MAPE(%) | MAE | RMSE | MAPE(%) |
| STGCN | 17.55 | 30.42 | 17.34 | 21.34 | 34.07 | 14.18 | 25.33 | 39.34 | 11.21 | 18.15 | 28.67 | 11.61 |
| STGCN# | **16.98** | **28.60** | **16.01** | **20.74** | **32.69** | **13.22** | **24.61** | **38.46** | **10.49** | **17.31** | **27.94** | **11.26** |
| DCRNN | 17.99 | 30.31 | 18.34 | 21.22 | 33.44 | 14.17 | 25.22 | 38.61 | 11.82 | 16.82 | 26.36 | 10.92 |
| DCRNN# | **17.21** | **28.97** | **17.49** | **20.37** | **32.14** | **13.28** | **23.99** | **37.02** | **11.18** | **15.93** | **25.24** | **10.10** |
| GraphWaveNet | 19.12 | 32.77 | 18.89 | 24.89 | 39.66 | 17.29 | 26.39 | 41.50 | 11.97 | 18.28 | 30.05 | 12.15 |
| GraphWaveNet# | **18.95** | **31.85** | **18.20** | **23.06** | **37.25** | **16.80** | **25.44** | **39.81** | **11.13** | **17.68** | **28.33** | **11.72** |
| ASTGCN | 17.34 | 29.56 | 17.21 | 22.93 | 35.22 | 16.56 | 24.01 | 37.87 | 10.73 | 18.25 | 28.06 | 11.64 |
| ASTGCN# | **16.77** | **28.91** | **16.80** | **20.51** | **32.84** | **14.00** | **22.79** | **36.53** | **10.22** | **17.84** | **26.38** | **11.02** |
| AGCRN | 15.98 | 28.25 | 15.23 | 19.74 | 32.16 | 13.14 | 22.37 | 36.55 | 9.12 | 16.18 | 25.65 | 10.30 |
| AGCRN# | **15.46** | **27.83** | **14.90** | **19.69** | **32.02** | **12.87** | **22.01** | **36.12** | **9.03** | **16.00** | **25.43** | **10.17** |
| STFGNN | 16.77 | 28.34 | 16.30 | 20.29 | 32.23 | 13.33 | 23.46 | 36.60 | 9.21 | 16.68 | 26.23 | 10.72 |
| STFGNN# | **16.33** | **27.96** | **16.11** | **19.97** | **31.82** | **13.01** | **23.10** | **36.44** | **9.20** | **16.29** | **25.94** | **10.36** |
| STG-NCDE | 15.57 | 27.09 | 15.06 | 19.21 | **31.09** | 12.76 | 20.53 | 33.84 | 8.80 | 15.45 | 24.81 | 9.92 |
| STG-NCDE# | **15.21** | **26.77** | **14.74** | **19.00** | 31.20 | **12.46** | 20.80 | <u>33.55</u> | 8.82 | **15.38** | <u>24.53</u> | **9.78** |
| ASTGNN | 14.80 | 24.81 | 14.89 | 18.93 | 31.36 | 12.52 | 20.03 | 33.43 | <u>8.41</u> | 14.94 | 25.08 | 9.81 |
| ASTGNN# | <u>**14.21**</u> | <u>**24.77**</u> | <u>**13.74**</u> | <u>**18.40**</u> | <u>**30.20**</u> | <u>**11.46**</u> | <u>**19.80**</u> | 33.55 | 8.92 | <u>**14.88**</u> | **24.69** | <u>**9.38**</u> |

## 5.4 Experiments on spatiotemporal graph learning

**Main experiments:** To evaluate the effectiveness of GFS, we incorporate it with those graph convolution backbones by replacing their embedded graph convolution component. Specifically, we use those backbones and their GFS variants to predict the urban traffics during the next hour with the traffics during the previous hour, and the average result over the next 12 prediction steps is shown in Table 4. Notice that # denotes that GFS is integrated with corresponding backbones. The results of each individual backbone and its corresponding GFS variant are grouped by double line for clearer comparison. Regarding each group, the better one is marked in bold, and the best performance over all models is highlighted with underlines. As demonstrated in Table 4, in most cases, GFS based variants outperform the corresponding backbones, and this indicates the superiority of our proposed GFS module in capturing spatial correlations. And the performances of the GFS variant based on ASTGNN are the best in most experiments, and utilizing GFS in STGCN, DCRNN, GraphWaveNet, ASTGCN, AGCRN, STFGNN, STG-NCDE, ASTGNN can respectively gain the improvements on all metrics by {4.35%, 4.92%, 4.24%, 5.63%, 1.39%, 1.61%, 0.92%, 2.33%} in average, and this indicates that our proposed GFS module is applicable to all existing graph convolution based spatiotemporal learning works. On the other hand, regarding all backbones, replacing their integrated graph convolution with GFS can gain the performance improvements by {2.92%, 2.63%, 3.95%} respectively on MAE, RMSE and MAPE. In summary, the results on spatiotemporal graph learning undoubtedly verify the effectiveness of GFS in spatial learning. For more fine-grained analysis, we further compare the step-wise performances of AGCRN, STFGNN, ASTGNN and their corresponding GFS variants for each individual prediction step on PeMSD4 and PeMSD8, and results are shown in Figure 6. The horizontal axis corresponds to different time steps and the vertical axis corresponds to the performances in different metrics and with different datasets. First, as observed, the performances of the GFS based variants are better than the performances of the corresponding backbones at almost all time steps, this verifies the superiority of GFS in capturing spatial correlations. Second, we discover that the performances of the GFS based variants decrease slower than the performances of the corresponding backbones with the increasing of time steps, this indicates that GFS module can significantly improve traditional spatiotemporal learning on multi-step predictions.

**Time consumption:** As analyzed, the time complexity of GFS is significantly better than that of graph convolution. In this part, we compare the experiment time consumption of GFS with that of graph convolution. Similarly, for each backbone, GFS is used to replace their graph convolution component, all models are trained and tested on PEMSD4 and PEMSD8, and the results are listed in Table 7. As shown, for all models, utilizing GFS can significantly reduce the time consumptions on both training and testing by 20% averagely. Considering that different backbones have diverse architectures, and such divergence may interfere with the experiments to a certain extent. For the sake of fairness, we construct a series of generated datasets with different numbers of vertices ranging from 100 to 2000 to further investigate the time consumption issue. Each dataset contains 1000 batches and each batch contains 64 samples. Based on these generated datasets, we test the time consumptions of single-layer GFS and single-layer GCN (using the definition in Equation 11), respectively. The results are shown in Figure 7a. As illustrated, the difference between the time consumptions of these two networks is relatively small while the number of vertices is small. However, since the time complexity of GCN is quadratic to the number of vertices, the time consumption of GCN increases much faster than that of GFS with the increasing of vertex number. Furthermore, We also test the time consumptions of AGCRN and its GFS based variant on the constructed datasets, and the results are shown in Figure 7b. As demonstrated, the time complexity of AGCRN# scales linearly with the increasing of vertex number, while the time complexity of original AGCRN increases quadratically with the increasing of vertex number. The experimental results on the generated dataset validate the temporal efficiency of GFS.

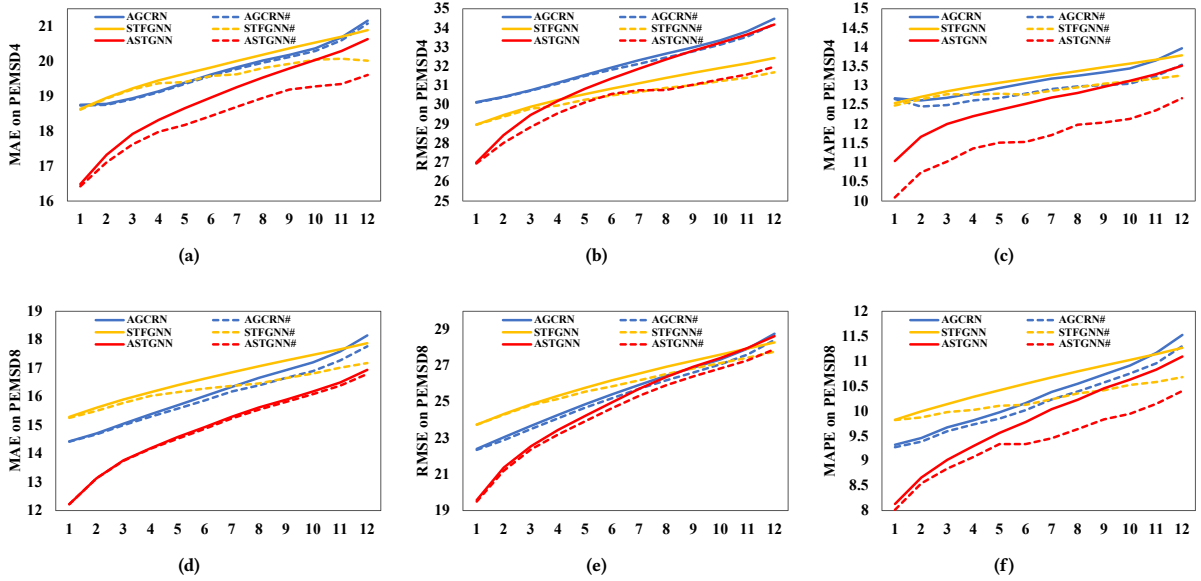

Figure 6: Step-wise performances of AGCRN, STFGNN, ASTGNN and their corresponding GFS variants on PeMSD4 and PeMSD8.

Table 5: Time consumption comparison between backbones and their GFS based variants on PEMSD4 and PEMSD8

| Model | PEMSD4 | | PEMSD8 | |
| --- | --- | --- | --- | --- |
| | Test | Train | Test | Train |
| STGCN | 2.89 | 22.38 | 1.62 | 13.48 |
| STGCN# | 2.76 | 18.05 | 1.57 | 10.48 |
| DCRNN | 3.64 | 25.39 | 2.53 | 23.94 |
| DCRNN# | 1.74 | 14.71 | 1.32 | 21.17 |
| GraphWaveNet | 1.70 | 28.31 | 1.66 | 23.57 |
| GraphWaveNet# | 1.65 | 24.96 | 1.56 | 19.04 |
| ASTGCN | 3.98 | 21.69 | 3.15 | 18.39 |
| ASTGCN# | 1.87 | 7.56 | 1.97 | 12.33 |
| AGCRN | 2.79 | 20.51 | 1.99 | 19.07 |
| AGCRN# | 1.17 | 13.79 | 1.14 | 14.90 |
| STFGNN | 9.56 | 103.79 | 7.82 | 88.62 |
| STFGNN# | 8.90 | 90.12 | 6.04 | 77.91 |
| STG-NCDE | 16.54 | 167.58 | 9.91 | 103.92 |
| STG-NCDE# | 13.90 | 142.69 | 9.12 | 95.67 |
| ASTGNN | 100.23 | 239.21 | 32.14 | 70.47 |
| ASTGNN# | 89.94 | 217.36 | 30.78 | 66.89 |

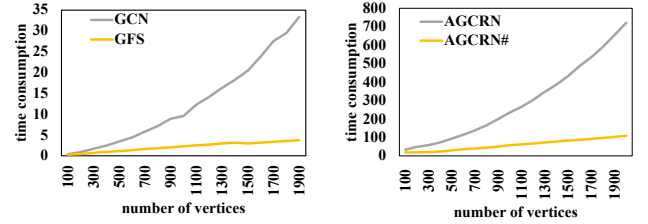

(a) Single-layer GFS and single-layer GCN

(b) AGCRN and its GFS variants

Figure 7: Time consumptions of Single-layer GFS and single-layer GCN with different numbers of vertices.

**Detailed Performances of state-of-the-art solutions on single vertex:** Regarding state-of-the-art solutions, ASTGNN and its corresponding GFS based variant ASTGNN#, we select two random vertices in one random day from the testing set, and evaluate the performances of these two state-of-the-art solutions for each individual time point during the selected two vertices, and the results are shown in Figure 8. Note that the data of vertices from about 17:30 to 20:30 is missed. As shown, both two models can achieve satisfying accuracies all the time. However, regarding some extreme scenarios including peak values and wild fluctuations, as has been highlighted with amplification rectangles, the performance curves of the GFS based variant can approximate the curves of truths more accurately, and this further illustrates the effectiveness of GFS.

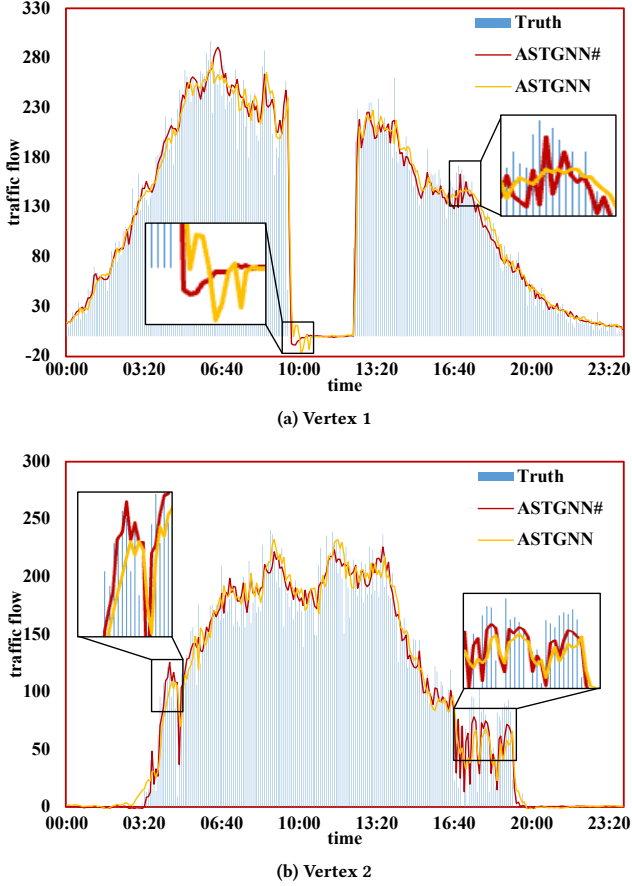

Figure 8: Detailed performances of state-of-the-art solutions at different vertices on PEMSD4.

### 5.5 Node classification on extreme large graph

**Main experiments:** In this subsection, we investigate the performance of GFS in processing extreme large graph, and a very simple three-layer stacked architecture is proposed in the experiments for both GFS and GCN. For training these two stacked networks, we randomly select training samples, train for 200 epochs, and test for 1000 rounds. The average classification accuracies are reported in Table 6. As observed, compared with GCN, utilizing GFS can improve the classification accuracy by 2.15% and 0.7% respectively on PubMed and Coauthor Physics. The results demonstrate the generalizability of GFS on general graph learning task and the scalability of GFS on large graph with massive vertices. Further, we also investigate the training accuracy of each epoch of GFS and GCN on two datasets, and the results are show in Figure 9. As can be easily observed, both two modules have similar convergence speeds and achieve satisfying fitting on Coauthor Physics. However, regarding PubMed, GFS is able to fit training samples better and thus has better representation capability than GCN. Such results witness the capability of GFS on modeling spatial correlations.

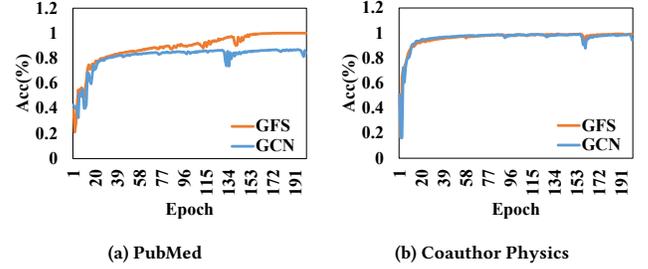

Figure 9: Training accuracy of each epoch of GFS and GCN.

**Time comsumption:** Regarding the previous node classification experiment, for each individual approach and dataset, we also record its time comsumption, and the results are also reported in Table 6. As shown, compared with GCN, GFS can reduce the time consumption by 1.64× and 9.6× respectively on on PubMed and Coauthor Physics. Compared with the previous spatiotemporal graph learning experiments on PEMSD3, PEMSD4, PEMSD7 and PEMSD8, our approach has more advantages over GCN in terms of time consumption with large graphs. The larger the graph is, the more advantage our GFS has in terms of time consumption, and this reflects that the great prospects of GFS in processing extreme large graph.

Table 6: Accuracy and time consumption of GFS and GCN.

| Dataset | Model | Acc(%) | Total time |
| --- | --- | --- | --- |
| PubMed | GFS | 0.8667343 | 26.429018 |
|  | GCN | 0.8452333 | 43.343687 |
| Coauthor Physics | GFS | 0.9670854 | 72.546249 |
|  | GCN | 0.9593931 | 698.05498 |

### 5.6 Investigation on graph-free architecture

In this section, to further verify the effectiveness of each individual component of the graph-free architecture, we carry out a series of ablative studies on PeMSD4, and the variants of GFS include:

- **Mean:** to evaluate the effect of layer normalization, we design this variant by replacing the layer normalization module of GFS with mean function, which is the equal to combine residual connection with a graph convolution with adjacency matrix as matrix $\mathfrak{R}$ defined in Equation10.
- **MeanP:** to further evaluate the performance gap between mean function and normalization operation, we design this variant by replacing the layer normalization of GFS with mean function but retains the affine parameters of layer normalization.
- **NoLNP:** to evaluate the effect of normalization operation itself, we design this variant by directly removing the layer normalization in GFS but keep the affine parameters of layer normalization in Equation 6, which equals to apply affine parameters directly on a graph convolution layer

Xu Wang[1], Pengfei Gu[1], Pengkun Wang[1], Binwu Wang[1], Zhengyang Zhou[1], Lei Bai[2*], Yang Wang[1*]

with adjacency matrix **I**. As affine parameters can be applied individually, we retain them in this variant.
- **NoRes:** to evaluate the effect of residual connection, the residual connection in GFS is removed in this variant.
- **LNNoP:** to evaluate the effect of affine parameters, we design this variant by removing the affine parameters of layer normalization.

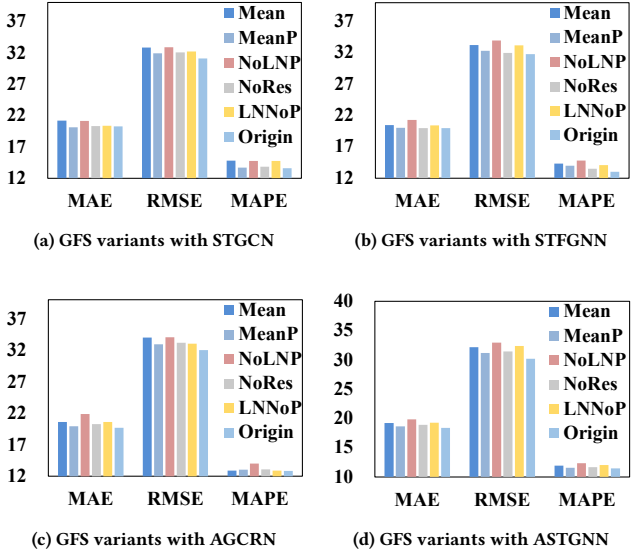

**Figure 10: Performances of different GFS variants with STGCN, STFGNN, AGCRN and ASTGNN on PEMSD4.**

To comprehensively evaluate the impacts of different components of GFS, we conduct a series of ablative experiments on PeMSD4 by incorporating all variants with the four representative approaches, i.e., STGCN, STFGNN, AGCRN, and ASTGNN, and the results are illustrated in Figure 10. As can be easily observed, GFS itself outperforms all variants with all backbones, and this first indicates that all components in GFS is effective to the final performances of GFS. And the variant of MeanP outperforms all other variants with all backbones except STFGNN, and this cross-verifies two points: i) the aggregation itself is more important than the way of aggregation, and ii) layer normalization is more effective than mean function on aggregating vertices. And the performances of NoLNP also indicates the first point which is consistent to the conclusion that we have obtained in Section 3. Comparing the performances of two solution pairs, MeanP vs. Mean and GFS vs. LNNoP, we discover that, no matter mean or layer normalization is used for aggregating vertices, affine parameters are vital and essential for GFS. Furthermore, based the performances of NoLNP and LNNoP, it is obvious that, the normalization operation itself, which introduces the aggregation of vertices, is more important than the mechanism of affine parameters, even though affine parameters can change the way that layer normalization aggregates vertices. This triple verifies that the aggregation itself is more important than the way of aggregation. And finally, the performance comparison between NoRes and GFS also verifies the importance and necessity of the residual connection component.

## 6 CONCLUSION AND DISCUSSION

**Conclusion:** In this paper, with extensive and deep-going experiments, we comprehensively analyze existing spatiotemporal graph learning models and reveal that extracting adjacency matrices with carefully design strategies, which are viewed as the key of enhancing performance on graph learning, are largely ineffective. Meanwhile, based on these experiments, we also discover that the aggregation itself is more important than the way that how vertices are aggregated. With these preliminary, a novel efficient GFS learning module based on layer normalization for capturing spatial correlations in spatiotemporal graph learning. The proposed GFS module can be easily plugged into existing models for replacing all graph convolution components. Rigorous theoretical proof demonstrates that the time complexity of GFS is significantly better than that of graph convolution operation. Extensive experiments verify the superiority of GFS in both the perspectives of efficiency and learning effect in processing graph-structured data especially extreme large scale graph data.

**Discussion:** In future works, there are some more interesting issues can be further discussed,

- The effectiveness of GFS further indicates that spending too much efforts on extracting adjacency matrix is the wrong region to spatiotemporal learning, and the reason why adjacency matrix is almost useless needs to be further explored. And Instead of relying on designing new adjacency matrix and incorporating it with graph convolution, how to effectively capture spatial correlations from spatiotemporal data need to further thought.
- Even though GFS has achieved promising performances on spatiotemporal graph learning, its performance is largely owed to the affine parameters of layer normalization. Considering that the shape of such parameters are predefined, the scalability of GFS is largely limited since GFS is not applicable to the scenario where the number of vertices is dynamic.

Graph-Free Learning in Graph-Structured Data: A More Efficient and Accurate Spatiotemporal Learning Perspective